*Joss Moorkens and Monja Jannet. Translation and data ethics. To appear in: S. Castilho, D. Kenny, J. Moorkens & M. I. Rivas Ginel (eds.). Developing AI literacy for translation Berlin: Language Science Press*

# Translation and data ethics

Joss Moorkens and Monja Jannet

*This chapter introduces data ethics for translation and discusses a number of ethical issues regarding data. These include issues of copyright, ownership and bias, which apply to all types of data, whereas there are further issues particular to minoritised language communities and those on either side of the digital divide, as data is implicated in power imbalances and dynamics. We close with a discussion of responsibility and regulation, and reflect on the growing attention on data ethics from translation studies researchers.*

- *What is data ethics?*
- *What are the most common ethical issues regarding translation data?*
- *How does data encode power dynamics?*
- *Will regulation solve these issues?*
- *What do translation researchers think about data ethics?*

## 1 Introduction

The metaphor of data as oil (Humby 2006), a natural resource just waiting to be collected and processed to give it value, is now a well-known trope. It serves an important purpose in separating the data, in our minds, from the creators of that data. It sounds objective. The data does not lie. As illustrated in the preceding chapters, data has acquired huge importance in translation. Data-driven machine translation (MT) has progressed through a succession of paradigms, from example-based to statistical to neural MT to large language models (LLMs), each more powerful than the last. As Kenny (2011: 7) wrote about statistical MT, the metaphors likening MT training data to food and natural resources to be 'harvested' made the very human data source -- translators -- invisible, "portraying translation as an undifferentiated mass, in which there is little scope to discern the particularities of individual translations, or traces of the work of individual translators".

A key change from statistical to neural MT, enabling a further degree of abstraction, was from symbolic to subsymbolic artificial intelligence (AI), whereby steps that were complex but comprehensible and explainable were replaced by machine learning processes that are a vast “stack of equations – a thicket of often hard-to-interpret operations on numbers” (Mitchell 2020: 24}. The ‘supervision’ of neural networks, following a path from source to target text for neural MT, has changed to (autonomous-sounding) self-supervision by LLMs that find patterns in data without explicit commands. The progressive automation of translation (among many other fields) means that human steps are fully or partially replaced by (now fully depersonalised) data-driven software. But data is not neutral, is not naturally occurring, and can be used to accrue wealth and power. This is why it is important to think about data ethics.

Ethics is the branch of philosophy that examines moral questions to evaluate whether an action is right or wrong, good or bad. There are several theories of ethics that may be applied to carry out evaluations in practical situations. Areas of applied ethics include business ethics, computer and information ethics, and, as data-driven software has taken precedence over computer hardware, data ethics. Although the term had been used before, Floridi and Taddeo (2016: 3) define data ethics as the subfield that examines

> moral problems related to data (including generation, recording, curation, processing, dissemination, sharing and use), algorithms (including artificial intelligence, artificial agents, machine learning and robots) and corresponding practices (including responsible innovation, programming, hacking and professional codes), in order to formulate and support morally good solutions (e.g. right conducts or right values).

Data ethics is particularly concerned with big data and machine learning, with translation as one of the main applications, highlighting the responsibilities of developers and users. Koskinen & Pokorn (2020: 3) define translation ethics as “the subfield that aims to understand what is good and bad, right and wrong in translatorial praxis”. To their contention that translation is at an intersection of languages and cultures, of creator and reader, we might also add the dynamic intersection of human and data-driven technology, necessitating a focus on translation data ethics.

In the following sections, we chart some ethical issues, discuss responsibility and regulation, and consider some future directions for research and discussion.

## 2 State of the art

If we follow Floridi & Taddeo (2016) from data to algorithm to practices, we can identify a number of ethical issues, which we expand on in this section.

### 2.1 Copyright, ownership and reuse

The Berne Convention from 1868 sets an international standard for copyright and ownership. There are some differences in how this is interpreted from one jurisdiction to another, but creative works are protected by copyright and translations as a derivative work that ‘shall be protected as original works without prejudice to the copyright in the original work’ \ World Intellectual Property Organization (WIPO) 1979: Article 2. The Berne Convention contains no definition of ‘original work’, and thus the concept is understood differently in different jurisdictions. In some European Union (EU) countries, this will depend on the extent of the authors’ intellectual creation, whereas in the United States, the Supreme Court precedent is to prove a ‘minimum degree of creativity’ Cabanellas 2015: 20). Troussel & Debussche (2014) argue that the level of originality of a translated work might provide grounds to claim copyright, but this has not been tested in courts.

The Berne Convention protects authors' moral right to be identified as the author of a work and to object when the presentation of that work may damage their reputation, and their economic right to benefit commercially by selling or licensing their work. Moorkens & Lewis (2019) add that the original author may not use the translation as the basis for a further translation without permission from the translator who, as translation moral rights owner, retains rights over the translation of a text even if it is out of copyright. There are further regulations and directives that may apply to translation, such as the European Database Directive, which allows 15 years’ protection for those who have made “a substantial investment [...] in obtaining, verifying and presenting the contents of the database” Troussel & Debussche (2014: 19).

Despite these potential routes to claim copyright, rights to translations (particularly specialised translations) have tended to be handed over to clients, either through precedent or contracts. Translators have usually not had a say in how their work was reused. This has become more complicated due to the scale of neural MT systems and LLMs, meaning that “tracing the intellectual contribution of a single translator, whose rights are unclear prior to training, is incredibly difficult or practically impossible” (Moorkens & Lewis 2019: 474). The increased focus on scale for LLM training means that bilingual parallel corpora are not enough for system training, and web crawling has become necessary for collecting data. As we mentioned in Chapter 2, LLM translation ability, rather than stemming from parallel training data, appears now to be an emergent ability from some form of internal alignment (Balashov 2025). Many countries have introduced copyright exception for LLM training, in an attempt to balance the protection of expression with stimulating innovation (Sag & Yu (2025) review exceptions in the UK, EU, Japan and Singapore).

## 2.2 Data bias

Biases are a risk of using AI systems that can lead to discrimination against individual users and certain groups. The IBM AI risk atlas (Bagehorn et al. 2025: 18) defines bias as “[occurring] when one group is unfairly advantaged over another due to decisions of the model. This might be caused by biases in the data and amplified as a result of the model’s training.” For instance, an AI-powered hiring system might unfairly disadvantage a certain demographic group, resulting in restricted economic opportunities for users pertaining to this group. Gender bias is a type of bias that is particularly concerning for automated translation. It can occur in translations from gender-neutral languages (for example, English) into gender-discriminatory languages such as German. Working on translations of this kind, translators need to choose a binary gender in many cases, such as for job positions. LLM gender decisions can be biased if they reflect stereotypes discriminating against a specific gender, for example, the assumption that doctors are usually male and nurses female.

Although LLMs tend to hallucinate, making up information while presenting it as facts, biases are typically not fabulated by the model but learned from the large textual corpora of training data. Hence, we can think of a biased LLM as a mirror showing us stereotypes present in certain societies; as Vallor (2024: 9) writes, “AI isn’t misaligned with our current values. It’s already expressing those values all too well.” The risk of biased training data and LLM output calls for more diversity in model development. Using a term borrowed from psychology, the current AI landscape is sometimes referred to as WEIRD, an acronym for western, educated, industrialised, rich and democratic, as the main characteristics of those developing and benefiting from LLMs.

To some extent, bias is a natural process of human psychology when our brains select what type and quantity of information is worth focusing on in a specific situation. Similarly, some AI-powered tasks may require a certain degree of differentiation between user groups. In order to avoid discrimination against users, it is important to establish fair criteria for AI-powered decisions. For instance, the decision as to whether to grant a loan should be based on the applicant's income (among other factors), rather than cultural or religious beliefs.

In addition to biases propagating societal stereotypes found in LLM training data, the use of synthetic data for training might also cause bias (See Chapter 2 describing synthetic data). Synthetic data is data that has been generated by an AI model itself. Including synthetic data into LLM training can be necessary, for example, if not enough authentic data is available in a certain domain. The increasing use of generative AI systems might lead to a future shortage of human-generated data ‘harvested’ from the internet, which is the main source of training data for LLMs. This development might force LLM providers to resort to synthetic data.

A best practice for LLM development employs different types of data sets in order to achieve viable output results after model deployment, comprising a training set (the largest part) and validation data. During the training phase, the latter is important to ensure that the model does not over-rely on a limited set of training data (a phenomenon known as overfitting). Finally, testing data is needed to evaluate the model’s performance on an unseen data set that was not included in the previously used sets. As Alemohammad et al. (2023) show, sampling bias can emerge if models are trained on synthetic data, whereby the quality and diversity of model output diminish when not enough new, previously unseen data is included in the training datasets.

In order to fight the various types of AI biases, both voluntary AI governance frameworks as well as legally binding AI regulations have introduced mitigation strategies. As described above, bias can be fuelled by an underrepresentation of certain groups' characteristics in AI training data. In this light, the EU AI Act (European Parliament and Council 2024: Article 10) requires an "examination [of training data] in view of possible biases that are likely to affect the health and safety of persons, have a negative impact on fundamental rights or lead to discrimination prohibited under Union law [...]". This examination procedure, however, is mandatory only for AI systems classified as high-risk under the EU AI Act. LLMs used for translation are currently not regarded as systems of high risk.

## 2.3 Minority languages in the digital domain

Minority languages are spoken by small communities of active speakers. Some minority languages are being actively used by so few speakers that they can be considered endangered languages. These languages might lose all active users in the future. Keeping in mind that the number of minority languages varies constantly and that it can be difficult to gain information about these communities, researchers estimate that, as of 2025, 44% of the world's languages are endangered (Eberhard et al. 2025).

Using LLMs to translate minority languages poses various challenges. Multilingual LLMs allow translation of various majority and minority languages. Although the models have been trained on predominantly English texts scraped from the internet, the unique capabilities of LLMs make it possible to generate content and translations into many other languages as well. Despite these multilingual capacities, Papadimitriou et al. (2023: 1194) found that translations into minor languages risk bias, whereby "grammatical structures in higher resource languages bleed into lower-resource languages". The scenario described by Moorkens et al. (2025: 225), in which an LLM, asked to produce text about breakfast food in Urdu, outputs text about bacon and eggs, not breakfast foods in Pakistan or India and potentially offensive to Urdu readers, is a likely outcome of an LLM trained on over 90% English language data, as was GPT-3 (Brown et al. 2020).

Moreover, the quality of LLM-generated translations (and LLM output in general) depends on the amount and quality of digitally available training data. Usually, minority languages (or low-resource languages, as they are sometimes referred to in the context of AI) do not possess large quantities of digital training data. Yet, it is not only the poor availability of training data that hampers the thriving of minority languages in the digital space. In order for speakers of minority languages to be able to benefit from high-quality language technologies, these languages need to achieve digital language equality, which can be defined as “the state of affairs in which all languages have the technological support and situational context necessary for them to continue to exist and to prosper as living languages in the digital age” (Gaspari et al. 2021: 4).

Here, technological factors comprise digital tools and services, datasets, and lexical resources such as grammatical information, among others. Contextual factors, on the contrary, refer to broader conditions of the European countries with factors such as economic, educational, legal and funding landscape. Analysing the European languages’ coverage with these digital resources, Gaspari et al. (2021) found strong variance between the individual languages, with English and German dominating the field for both technological and contextual factors. Small languages are at the lower end of the rankings.

In order to strengthen the digital availability of language technologies for minority languages, it would, hence, be necessary to equip the respective language communities not only with AI tools such as LLMs, as this only contributes to the technological factors. Instead, for small languages to benefit from high-quality LLM-powered translation, the communities would also need to have highly developed contextual factors.

As Paullada (2020: 4) notes, “Western discourses of language endangerment uncritically treat the development of technologies for low-resource languages as a social good”. Not all language communities want their data to be commoditised by external commercial technology developers. For example, Te Reo Māori communities have asserted collective ownership of linguistic data as part of the Māori Data Sovereignty network,2 which treats language data not merely as a resource, but as a collectively owned taonga or treasure. They assert that Māori should control the creation, collection, usage, and storage of their language data to prevent colonisation, exploitation, and misuse by big tech corporations and their AI systems. Theirs is one of many grassroots efforts to bridge the digital divide for language technology.

## 2.4 The global digital divide

The gap between majority and minority language communities in terms of representation in data, data ownership, and information asymmetry can lead to a disempowering feedback loop as biased algorithms perpetuate existing disadvantages. This is an example of the digital divide, originally conceptualised as the gap between those with access to information and communication technology and those with little or none, but now broadened to include technology skills, use and performance. The International Telecommunication Union (2025) esti-mates that 2.2 billion people, over 25% of the world's population, are entirely offline. These people are more likely to live in rural areas with socioeconomic disadvantages; only 14 per cent of rural residents in low-income countries are online (International Telecommunication Union 2025: 31).

Although many more people come online every year, this can bring new challenges, as these new internet users have not built up digital literacy skills through years of internet use, leaving them vulnerable to issues of cybersecurity and data privacy (Misra & Srivastava 2024). Dignum (2026) also notes the paradox that communities with little data interactivity are almost invisible to big tech companies, who therefore do not tend to cater for their needs. For example, migrant workers from Myanmar (a country with low connectivity rates) in Thailand struggled to use online MT due to problems with Myanmar keyboard input and unpredictable translation quality. This was exacerbated by their limited digital skills and lack of awareness of privacy issues regarding personal data and interactions with a free online service (Phanthaphoommee et al. 2026). We thus have a situation in which MT and technology users in the Global South are little represented in training data, yet tend to have more workers in data annotation than in wealthy countries, and the "unequal distribution of environmental harm affects the Global South and economically disadvantaged groups the most, despite their contribution to carbon emissions being the least" (Anand & Coeckelbergh 2026: 144).

## 3 Responsibility and regulation

Depending on the jurisdiction, there are different approaches to AI regulation. The EU's AI Act is the most prominent of a host of legislation, China's AI ethics guidelines provide some top-down regulation of development. In the US, the Trump administration has tended to leave AI governance to companies in an effort to stimulate innovation. Some big tech companies have teams of interdisciplinary experts to assess ethical concerns before deployment, such as Microsoft's AI Ethics Committee, but the sidelining of ethics experts (in Google, for example) suggests that these teams are not considered to be of primary importance. Ongoing legal cases may change this if organisations are forced to take responsibility for adverse outcomes from their data-driven AI products.

Drawing from Lee (2024: 8), we can think of responsibility in four different, but not exclusive, ways. There is role responsibility, in which individuals are responsible as part of their social role as, for example, a parent, an employee or a friend. Secondly, there is causal responsibility, whereby an individual has (or has not) done something, perhaps due to carelessness or ineptness, leading to a negative outcome or event. Thirdly, blameworthiness is similar, but involves an individual not following a long-standing norm or instruction. Fourthly, there is liability, when the individual who was responsible in one of the above ways is legally obliged to pay compensation.

Attributing responsibility or liability regarding data-driven systems can be very difficult. The scale and complexity of big data-driven systems such as LLMs make accountability difficult to attribute, as causation within a 'black box' is difficult to prove. The problem of many hands means that so many people have been involved in the preparation of a whole that it becomes impossible to identify a single culprit to blame for problems. We might expect that the producer company should still bear responsibility, but large companies are careful to avoid this. Then we have the problem of responsibility when an individual creates something with the assistance of AI. It is not controversial to propose that an individual should bear responsibility for AI-assisted harmful or misleading material, even if the AI output is unchanged. This assumes that the individual can actually change the output, unlike greyed-out text in a translation interface. However, what about responsibility for positive outcomes? Porsdam Mann et al. (2023) propose that positive outcomes are likely to be credited to AI in workplaces where the role of humans is to oversee and repair AI output. They propose detailed and transparent disclosure of roles to rebalance what they call a credit-blame asymmetry.

Due to the fast pace of technological development, it has been difficult for legislation to keep up, with additional pressures not to constrain innovation. The existing laws, such as from the final report of the US National Commission on New Technological Uses of Copyrighted Works (National Commission on New Technological Uses of Copyrighted Works 1979) contend that the "author is the one who employs the computer" and thus that they, rather than the machine, should have liability in case of harm. More recently, within the EU, the AI Liability Directive (European Commission 2022) was proposed in 2022 to reduce the burden of proof of causation of harm for litigants. However, the proposed Directive was withdrawn as of the 2025 Work Programme.3 The most recent legislation does not necessarily attribute liability, but instead places obligations on AI developers and providers.

Certain types of data are specifically protected by EU data protection law. The General Data Protection Regulation (GDPR) watches over personal data that is "any information relating to an identified or identifiable natural person" (European Parliament & Council of the European Union 2016: Article 4). Examples for personal data listed in the GDPR include "name, an identification number, location data, an online identifier" as well as more broad factors such as data describing "the physical, physiological, genetic, mental, economic, cultural or social identity" (European Parliament & Council of the European Union 2016). The GDPR applies to various areas of data processing, and AI might fall into its scope. In theory, LLMs can be built without including any personal data in the training data. Yet, web scraping is a common practice for data gathering in LLM development. Hereby, AI developers automatically search the web for large amounts of textual data needed at various stages of the AI life cycle, from model training to testing and performance evaluation. During this data collection process, personal data might enter the LLM development data. Moreover, personal information may come into play when users themselves include it in prompts written to instruct LLMs for a specific task, for instance, when personal information is part of a text to be LLM-translated.

Processing personal data in AI applications poses challenges in light of the GDPR. The regulation generally prohibits making automated decisions about individuals in cases where such an automated decision has "legal effects concerning him or her or similarly significantly affects him or her" (European Parliament & Council of the European Union 2016: Article 22). This requirement might be in conflict with using LLMs, as concerned individuals' (also called data subjects) rights are difficult to enforce due to the black box character of LLMs. According to the GDPR, data subjects have the right to opt out of personal data usage, to correct any information included in personal data and to register a complaint about how a company is processing their personal data. In addition, personal data processing is allowed only after getting the data subject's explicit consent allowing to do so. All these requirements are difficult to adhere to when working with AI.

In order to better protect the privacy rights of users and data subjects whose personal data might have been collected during the web scraping process, privacy-enhancing techniques can be used. Pseudonymisation and anonymisation are privacy-enhancing techniques whereby identifiable information is modified (perhaps by replacing it with a pseudonym) or removed completely from the data. Data edited so that the subject cannot be recognised is no longer considered personal data under GDPR. However, the requirements for completely anonymising personal data are so high that this technique is not practical in many cases. The discourse remains open as to whether or not we might have to reconcile with the usage of personal data in AI applications. Including personal data in LLM training data might be useful for certain areas of research, such as the topic of bias, where gender, age, geographic origin and similar information might be needed to analyse an LLM for biases discriminating against members of certain demographic groups. Synthetic data that contains patterns of bias are often used for these purposes.

## 4 Future directions

As noted in Chapter 2, the trend is towards ever-larger language models without consideration of data ethics. The combination of AI hype and political pressure (see Sweney 2025) is making regulation difficult. Harms related to data are very difficult to quantify, but it is important for educators to understand them as we build our own AI literacy. Translation researchers increasingly address data ethics, with some recent approaches summarised in this section.

Critical translation technology researchers have borrowed from sociology and philosophy of technology to describe ethical issues regarding translation and data. For example, Olohan (2017) uses Feenberg's critical theory to explain how translation technology encodes and reinforces hegemonic values. Baumgarten (2025: 64), in the Routledge Handbook of Translation Technology and Society, diagnoses this as data colonialism using translation technologies that "owe their existence to industrial extractivism, and they are subject to the expropriation, control, and monetisation of their linguistic assets by the biggest tech players". The chapter by Mellinger & Hanson (2025) in the same handbook is one of the few long-form translation studies publications on data and ethics. The focus is on data as a source of optimisation, charting employers' use of data-driven technologies to streamline processes. They note how data are socially constructed based on the collection process and can only be an incomplete representation of a population. Their predictions of tensions between an "increasing emphasis on quantification and optimisation", particularly from businesses that want to maintain data confidentiality and a competitive advantage, and researchers' "greater reflection on open science, data management, and ethics" sound highly likely (Mellinger & Hanson 2025: 533).

Reijers & Dupont (2023) explicitly attempt to bridge the gap between MT and philosophy of technology, suggesting that approaches from the latter could help us to understand ethical issues with the former. They argue that "technologies are not mere instruments; they are not neutral with regards to human actions and values but instead mediate reality", that "technologies can promote or prohibit human action and delegate responsibility", and that there is interaction between our social and political norms and technological design (Reijers & Dupont 2023: 13). Nunes Vieira (2023) follows this call with a postphenomenological analysis of MT-mediated interactions that have the potential to empower and misinform. Moorkens (2025) also employs postphenomenology to examine how MT magnifies the input of technologists while reducing the perceived human input, concealing the role of data annotators, and constraining output to the most commonly-occurring words in training data from a specific time period in the past, all of which exacerbates risk, particularly for low-resource language users.

## 5 Key ideas and takeaways

Following the discussion of data types and collection in Chapter 2, this chapter introduces a number of ethical issues regarding data and the applied field of data ethics. Issues of copyright, ownership and bias apply to all types of data, whereas there are further issues particular to minoritised language communities and those on either side of the digital divide, as data is implicated in power imbalances and dynamics. We close with a discussion of responsibility and regulation, and reflect on the growing attention on data ethics from translation studies researchers.

## 6 Further reading

- Floridi and Taddeo (2016) is the best place to begin your introduction to data ethics.
- Mellinger and Hanson's 2025 discussion of data issues around the translation industry's data-driven optimisation and its risks is a great contribution.